\pdfoutput=1
\documentclass[runningheads]{llncs}
\usepackage[final,year=2026]{accv}

\usepackage[T1]{fontenc}
\usepackage[utf8]{inputenc}
\usepackage{graphicx}
\usepackage{booktabs}
\usepackage{amsmath}
\usepackage{siunitx}
\usepackage{xcolor}
\usepackage{tikz}
\usetikzlibrary{arrows.meta}
\usepackage{microtype}
\usepackage[breaklinks,colorlinks,citecolor=blue]{hyperref}

\begin{document}

\title{The Text Beside the Image: Detection, Utility and Leakage
  for Trustworthy Multimodal Medical Data and Beyond}

\author{%
  Andreas Maier\inst{1}\and
  Monica Hinrichs-Mayer\inst{2}\and
  Franziska Weber\inst{2}\and
  Niklas Lackner\inst{3}\and
  Matthias May\inst{3}\and
  Bernhard Kainz\inst{2,4}\and
  Siming Bayer\inst{2}}

\authorrunning{A. Maier et al.}
\titlerunning{The Text Beside the Image}

\institute{%
  Pattern Recognition Lab, Friedrich-Alexander-Universit\"at Erlangen-N\"urnberg, Germany\and
  Friedrich-Alexander-Universit\"at Erlangen-N\"urnberg, Germany\and
  Universit\"atsklinikum Erlangen, Germany\and
  Imperial College London, United Kingdom}

\maketitle

\begin{abstract}
Medical images are released with the reports that describe them, and protecting the image does
not protect the report. This paper measures the text component of such releases. We measure identifier detection, downstream utility and residual identity
leakage on the same documents, with the pseudonymisation policy as the variable under test: 15
detectors, three release conditions and four corpora of medical reports, legal judgments, news and
other genres, and e-mail, in German, English, Chinese and Arabic. A fixed 13-detector union
reaches a person sensitivity of \num{0.9998} at specificity \num{0.8686} on the medical reports,
\num{0.9958} at \num{0.8504} on the legal judgments, \num{0.9352} at \num{0.9318} on news and
other genres, and \num{0.9906} at \num{0.6235} on e-mail. With this ensemble, frequency matching
with a public name list recovers zero identities by alignment across the four corpora; the names it
got right were ones the detector missed, left in clear text. Cross-document linkage ranks the
correct person first for \SI{0.93}{\percent} of e-mail queries without training and
\SI{3.94}{\percent} with it, against $1/3697$ chance and \SI{71.98}{\percent} on unmodified text.
On the medical reports it recovers nothing without training and \SI{0.71}{\percent} of 138 queries
with it, against $1/207$ chance and a \SI{2.73}{\percent} ceiling on unmodified text.
\end{abstract}

\begin{keywords}
pseudonymisation \and clinical text \and text in multimodal releases \and privacy evaluation
\end{keywords}

\section{Introduction}\label{sec:introduction}

Clinical data are multimodal: a chest radiograph travels with its report, a cardiology study with
its discharge letter. Releasing either half requires both to be protected, since an image may need
defacing or removal of burned-in text while its report needs identifier replacement. Artificial
intelligence scales both tasks, and its errors vary by distribution, language, script and identifier
class. Packh\"auser et al.~\cite{packhauser2022reid} re-identified patients from de-identified chest
radiographs and priced a defence against diagnostic utility~\cite{packhauser2023anonymization};
pathological speech has been evaluated against privacy, utility and
fairness~\cite{arasteh2024speaker}; clinical text is instead kept with the model inside the
hospital~\cite{hou2025llama}. We ask what survives transformation of the text, and what protecting
it costs.

\emph{De-identification} reduces identifying information. Reversible \emph{pseudonymisation}
replaces an identifier with a realistic \emph{surrogate}; \emph{anonymisation} protects individuals. 
Stable pseudonyms preserve longitudinal utility but hand an adversary a persistent handle.
European guidance regards keyed message authentication codes as
robust~\cite{enisa2021pseudonymisation}, and the health-informatics standard is revised
accordingly~\cite{din25237}.

Detection benchmarks ask whether an identifying span was
found~\cite{vats2026redact,jha2026piibench,uppala2026privacyfilter}; utility studies measure
downstream change~\cite{vakili2022utility,berg2020impact} or information
loss~\cite{trienes2024infolossqa}; re-identification studies measure attack success. Frameworks
combine technical performance, information loss and intrusion tests~\cite{mozes2021nointruder},
while surrogate systems and masking benchmarks cover single
pieces~\cite{eder2019emails,pilan2022tab,osborne2022bratsynthetic}. No study states together what one
policy buys and what it costs. Data-protection officers release documents and cases, not average
tokens, as one unchanged mention can identify someone after thousands were removed, so we measure
surviving findings, exposed cases and linkable entities rather than token rates alone.

Our design covers four corpora in German, English, Chinese and Arabic: clinical letters, legal
judgments, news and other genres, and e-mail. Every document receives the same detector comparison,
release conditions and measurements. Two predictions are under investigation: that detector precision
governs utility more strongly than the choice between surrogates and typed placeholders, and that
detection sensitivity moves the two attacks in opposite directions, because removing a name obstructs
context linkage while leaving a larger, more stable population for frequency matching. 

\section{Material and Methods}\label{sec:methods}

\subsection{Corpora}\label{sec:corpora}

Reference provenance differs across the corpora, and with it the detection task
(Table~\ref{tab:corpora}). CARDIO:DE is the clinical anchor. To use it here, we inserted gold
identifiers into an already de-identified release, which makes its name distribution synthetic. TAB
manually annotates identifiers and document-scoped co-reference in court
judgments~\cite{pilan2022tab}; a \emph{named entity} is a predefined person, organisation, location
or other type, its occurrences form a co-reference chain. OntoNotes covers three languages but was
annotated for linguistics rather than privacy. Enron derives genuine cross-document identity from
message headers but contains exposed personal information~\cite{noever2020enron}.

\begin{table}[t]
  \caption{The corpora separate domain, language, and where the reference annotation came from.}\label{tab:corpora}
  \centering\small
  \begin{tabular}{@{}l@{\hspace{1.2em}} S[table-format=5.0]@{\hspace{1.2em}} S[table-format=8.0]@{\hspace{1.2em}} S[table-format=2.1]@{\hspace{1.2em}} l@{\hspace{1.2em}} l@{}}
    \toprule
    Corpus & {Docs} & {Tokens} & {Ment./doc} & Lang. & Reference annotation \\
    \midrule
    CARDIO:DE & 400 & 885059 & 78.3 & de & constructed \\
    TAB & 1268 & 1844804 & 81.2 & en & manual \\
    OntoNotes & 5994 & 4726594 & 45.5 & en/zh/ar & linguistic \\
    Enron & 58636 & 15507925 & 12.2 & en & header-derived \\
    \bottomrule
  \end{tabular}
\end{table}

Therefore CARDIO:DE ground truth covers only inserted identifiers, and \SI{84.7}{\percent}
of its gold tokens are dates, which the detectors do find but the conditions pass through by design.
Sensitivity, replacement, exposure and the attacks are therefore all scored over the types actually
replaced, and the retained types are not part of our evaluation. Retained dates are themselves a
re-identification channel, easy to spot, and a release keeping them needs date shifting. A TAB applicant is
usually named once but has a median of 17 later role mentions, and only Enron has natural
cross-document identity.

\subsection{Conditions}\label{sec:conditions}

Condition A is unmodified text. In condition B, each detected identifier becomes a realistic \emph{surrogate},
such as another name, stable across the corpus, indexed by HMAC-SHA256, a keyed one-way digest
irreproducible without the secret. Condition C uses an unindexed placeholder such as
\texttt{[PERSON]}. B and C differ only in replacement form, following European guidance and prior
e-mail pseudonymisation~\cite{enisa2021pseudonymisation,eder2019emails}, and replace identical
spans.

\begin{figure}[t]
  \centering
  \resizebox{\textwidth}{!}{%
  \begin{tikzpicture}[
      x=1mm,y=1mm,
      box/.style={draw=blue!55!black, fill=blue!6, rounded corners=1.2mm,
                  line width=.45pt, align=center, inner sep=1.5mm, font=\scriptsize},
      measure/.style={box, draw=orange!70!black, fill=orange!8,
                      minimum width=27mm, minimum height=7mm},
      line/.style={align=left, font=\scriptsize, inner sep=.5mm},
      tag/.style={font=\scriptsize\itshape, text=black!55, inner sep=.5mm},
      hit/.style={fill=teal!16, rounded corners=.6mm, inner sep=.5mm},
      kept/.style={fill=red!12, rounded corners=.6mm, inner sep=.5mm},
      flow/.style={-{Latex[length=1.5mm]}, line width=.45pt, draw=black!70},
      rail/.style={line width=.45pt, draw=black!70},
      rewrite/.style={-{Latex[length=1.6mm]}, line width=.5pt, draw=teal!55!black, dashed}
    ]
    \node[box, minimum width=37mm, minimum height=15mm] (det) at (18.5,31)
      {detector pool\\\scriptsize 15 rules, encoders, LLMs\\[-.3mm]
       \scriptsize union / vote / intersection};

    \draw[flow] (det.east) -- (41,31);
    \draw[rail] (41,22) -- (41,40);
    \foreach \y in {40,31,22} { \draw[flow] (41,\y) -- (44.5,\y); }

    \node[tag, anchor=east] at (50,40) {A};
    \node[line, anchor=west] at (51.5,40)
      {Frau \tikz[baseline]{\node[hit,anchor=base]{Anna Weber};}, geb.\
       \tikz[baseline]{\node[kept,anchor=base]{12.03.1948};},
       \tikz[baseline]{\node[hit,anchor=base]{Klinik Nord};}.};

    \node[tag, anchor=east] at (50,31) {B};
    \node[line, anchor=west] at (51.5,31)
      {Frau \tikz[baseline]{\node[hit,anchor=base]{Keller};}, geb.\
       \tikz[baseline]{\node[kept,anchor=base]{12.03.1948};},
       \tikz[baseline]{\node[hit,anchor=base]{Klinik S\"ud};}.};

    \node[tag, anchor=east] at (50,22) {C};
    \node[line, anchor=west] at (51.5,22)
      {Frau \tikz[baseline]{\node[hit,anchor=base]{\texttt{[PERSON]}};}, geb.\
       \tikz[baseline]{\node[kept,anchor=base]{12.03.1948};},
       \tikz[baseline]{\node[hit,anchor=base]{\texttt{[ORG]}};}.};

    \draw[rewrite] (83,28.2) -- (83,25.3);
    \node[tag, anchor=north, text=teal!45!black] at (83,17.6)
      {identical spans, so B rewrites to C};

    \node[measure] (m1) at (131,40) {detection\\and exposure};
    \node[measure] (m2) at (131,31) {downstream\\utility};
    \node[measure] (m3) at (131,22) {attacks and\\stability};
    \foreach \y in {40,31,22} { \draw[flow] (116,\y) -- (117.5,\y); }
  \end{tikzpicture}%
  }
  \caption{One set of detector spans generates three release conditions, on an invented header line
  in the style of a discharge letter. Detected identifiers (teal) become surrogates in B and typed
  placeholders in C; dates (red) pass through by design. A surrogate replaces a whole detected span
  with a single name, so the two-token name becomes one. B and C replace the same spans, so C is a
  rewrite of B. Every condition meets every measurement family on the same documents.}
  \label{fig:overview}
\end{figure}
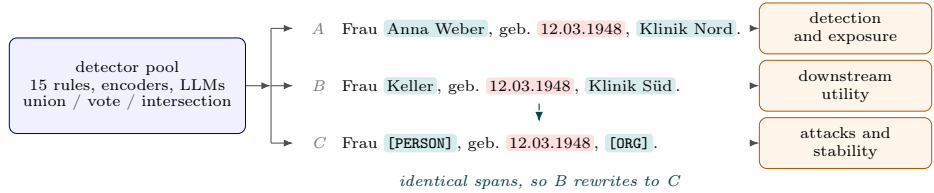

Before hashing, the key normaliser casefolds mentions, collapses whitespace and removes titles and
punctuation, so \emph{Dr. Weber} matches \emph{weber} but \emph{Weber} does not match \emph{Kay
Weber}. Unrelated people may therefore share a key while one person's initials and full name may
not; Sect.~\ref{sec:results-stability} reports both failures. Figure~\ref{fig:overview} evaluates
every condition on identical documents and measurements.

\subsection{Detectors, Ensembles and Operating Points}\label{sec:detectors}

The 15 frozen detectors comprise a rule engine~\cite{presidio}, a German domain
tagger~\cite{akbik2019flair,eder2019emails}, two zero-shot models~\cite{zaratiana2024gliner}, a
multilingual {NER} encoder~\cite{conneau2020xlmr}, two clinical
de-identifiers~\cite{chambon2023deid,johnson2020deid} and eight large language
models~\cite{openai2025gptoss,gemma2026gemma4,microsoft2025phi4mini,mistral2025magistral,qwen2025qwen3,granite41,mistralsmall32};
the repository, we release with acceptance of the paper, lists every model, version
and prompt. Cached spans are combined by union, intersection or at-least-$k$ vote. Two rules recur
below: the \emph{permissive} rule is the union of all 15, the \emph{precise} rule a majority, 8 of
the 15. Leakage uses the configuration we recommend, a fixed 13-detector union; two large language
models were excluded from it, because preliminary runs identified them as one to
two orders of magnitude slower per document without improving detection. 

To explore ensembling, each corpus reports fast and maximum detector triplets for sensitivity, the share of replaced-type
identifier tokens detected, and specificity, the share of all other tokens left untouched; for each parallel run we report slowest member of the ensemble. 
Our search covers singles, pairs and triples. Still for maximum performance, we recommended the 13-detector ensemble.
Note that it does exceed the maximum of Table~\ref{tab:operating-points} most of the time.

\subsection{Measurements}\label{sec:measurements}

Detection is scored by token and by entity, an entity counting as protected only when every one of
its mentions was replaced. Exposure uses document, case and entity denominators, because few leaked
tokens can expose many people. Frozen models analyse all three conditions for named-entity agreement, medication extraction
and section classification; agreement is span F1 between the outputs on original and transformed
text, and the original-text score accompanies every comparison. Continuous pairs use the Wilcoxon
signed-rank test, binary correctness McNemar's test, and Benjamini--Hochberg correction applies
within task families. Utility and attack figures carry one standard deviation over documents,
detector combinations or folds; detection and exposure are exact population proportions.

A \emph{collision} gives distinct people one surrogate, \emph{fragmentation} gives one person
several within a document, and \emph{drift} changes it between documents. These apply to B, which
has a mapping, and not to placeholders, which merge every entity of one type by design.

\subsection{The Attacks}\label{sec:attacks}

\emph{Frequency matching} aligns surrogate frequencies with real-name frequencies. Published attacks
on pseudonymised free text target the identifiers a system missed rather than the surrogate
distribution it produced, as the parrot attack does~\cite{carrell2019parrot}, so we transfer the
alignment model from attacks on encoded identifiers in record linkage and disclosure-risk
work~\cite{vidanage2022survey,naveed2015inference}. Two priors run over one observation, following
Kuzu et al., Niedermeyer et al. and Vatsalan et
al.~\cite{kuzu2013practical,niedermeyer2014cryptanalysis,vatsalan2014taxonomy}: a corpus-internal
prior granting perfect knowledge, an oracle upper bound, and a public name-frequency list, the
realistic adversary.

The attacker runs its own detector on the released text, so it can identify names that the defender failed to replace. All detected surfaces remain in the evaluation denominator, including ambiguous and unresolved cases~\cite{dankar2012estimating}. If the same surrogate represents $k$ different entities, a correct recovery receives a score of $1/k$ rather than being excluded. The attacker abstains when the score margin is too small, using $(\max-\max_2)/\sigma < \varphi$, where $\max$ and $\max_2$ are the best and second-best candidate scores, $\sigma$ their spread across candidates, and $\varphi$ the abstention threshold, all computed from its own candidate scores~\cite{narayanan2008robust}. We therefore report both the fraction of queries answered and $c@1$, which rewards correct answers while allowing justified abstention~\cite{penas2011simple}, together with the chance level implied by the candidate set~\cite{rocher2019estimating}. Realistic surrogates are intended to make residual identifiers difficult to distinguish from replaced ones~\cite{carrell2013hiding}; the public-prior attack tests this property directly. Varying the prior's size and publication date then quantifies the value of additional attacker knowledge~\cite{bindschaedler2018cryptanalysis}.

\emph{Context linkage} ranks identities by cosine similarity between surrounding-word profiles;
\emph{learned linkage} fits the same task on labelled, entity-disjoint examples. Both use identical
document-disjoint folds, and only the labels of held-out queries are hidden: those people stay in
the complete candidate list with every distractor, so neither attack is told that a query has a
match, and their difference measures learning rather than gallery size. Cross-document identity
permits this on Enron and CARDIO:DE but not on TAB or OntoNotes.

\emph{LLM candidate ranking} gives a language model (\emph{gpt-oss-120B}) one marked mention and ten candidate names, with
or without auxiliary text about each candidate from the released condition. Because conditions B and C replace
identical spans, B can be rewritten into C, so we report this attack on B as presented, on B after
that rewrite, and against a ranker told the scheme. The one-in-ten closed world measures an attacker
holding a shortlist, not population re-identification. A \emph{dictionary lookup} would hash
candidate names; it is defined but not run, because no condition here is unkeyed.

\section{Results}\label{sec:results}

\subsection{Detection}\label{sec:results-detection}

\begin{table}[t]
  \caption{Operating points on both error rates. Sensitivity is over the identifier types the
  conditions replace, specificity over every other token. Cost is seconds per document; the rule is
  $\cup$ union, $\cap$ intersection, $v$ at-least-$k$ vote, $s$ a single
  detector.}\label{tab:operating-points}
  \centering\scriptsize\setlength{\tabcolsep}{2.2pt}
  \resizebox{\textwidth}{!}{%
  \begin{tabular}{@{}l *{4}{S[table-format=1.4] S[table-format=1.4] r}@{}}
    \toprule
    & \multicolumn{3}{c}{CARDIO:DE} & \multicolumn{3}{c}{TAB}
    & \multicolumn{3}{c}{OntoNotes} & \multicolumn{3}{c}{Enron} \\
    \cmidrule(lr){2-4}\cmidrule(lr){5-7}\cmidrule(lr){8-10}\cmidrule(lr){11-13}
    Operating point & {Sens.} & {Spec.} & Cost & {Sens.} & {Spec.} & Cost
                    & {Sens.} & {Spec.} & Cost & {Sens.} & {Spec.} & Cost \\
    \midrule
    max.\ sensitivity & 0.9919 & 0.9670 & $57.633^{\cup}$ & 0.9253 & 0.8986 & $14.210^{\cup}$
                      & 0.7466 & 0.9763 & $9.450^{\cup}$  & 0.9968 & 0.7745 & $25.557^{\cup}$ \\
    fast sensitivity  & 0.8947 & 0.9769 & $0.301^{\cup}$  & 0.8926 & 0.9242 & $0.219^{\cup}$
                      & 0.6843 & 0.9731 & $5.376^{\cup}$  & 0.9222 & 0.7420 & $0.057^{\cup}$ \\
    max.\ specificity & 0.5230 & 0.9997 & $116.875^{v}$   & 0.5014 & 0.9959 & $98.110^{v}$
                      & 0.5034 & 0.9960 & $0.103^{\cup}$  & 0.5292 & 0.9448 & $20.931^{\cap}$ \\
    fast specificity  & 0.7807 & 0.9862 & $0.263^{s}$     & 0.5198 & 0.9934 & $0.117^{s}$
                      & 0.5034 & 0.9960 & $0.103^{\cup}$  & 0.6920 & 0.8781 & $0.026^{s}$ \\
    \bottomrule
  \end{tabular}}
\end{table}

Table~\ref{tab:operating-points} gives the cost--quality front for our detector triplet ensembles, with sensitivity the primary criterion for protection. Maximum-sensitivity configurations reach 0.9919 on CARDIO, 0.9253 on TAB, 0.7466 on OntoNotes, and 0.9968 on Enron. The faster alternatives retain 0.8947, 0.8926, 0.6843, and 0.9222 sensitivity, respectively, while sharply reducing cost; for example, CARDIO falls from \SI{57.633}{s} to \SI{0.301}{s} and Enron from \SI{25.557}{s} to \SI{0.057}{s} per document. The OntoNotes specificity rows coincide because the \SI{10}{\percent} band admits all 414 candidates.

Within OntoNotes, language effects depend on what is detected. Under the recommended ensemble,
sensitivity is 0.9890 on English, 0.8680 on Chinese and 0.6137 on Arabic; pooled sensitivity is
0.8011.
Arabic supplies \SI{45}{\percent} of identifier tokens from \SI{7}{\percent} of documents, so it
dominates the pooled result while remaining the hardest case. Person sensitivity transfers better
at 0.9913, 0.9315 and 0.8783, pooling to 0.9352. The thirty-seven-point spread therefore lies in the
other identifier classes rather than the people targeted by the attacks.

\subsection{Utility}\label{sec:results-utility}

At the high-sensitivity operating point, utility depends strongly on replacement form. On CARDIO, named-entity agreement is $0.611\pm0.082$ with surrogates but only $0.004\pm0.008$ with placeholders. The latter largely reflects the frozen recognizer rather than semantic degradation: it never tags \texttt{[PERSON]} as a name, so every replaced mention is counted as a disagreement. Surrogates therefore preserve token-level named-entity utility substantially better than placeholders. The same ordering holds on the other three corpora: named-entity agreement under the permissive rule is 0.478 on TAB, 0.295 on OntoNotes and 0.330 on Enron, rising to 0.898, 0.888 and 0.754 under the precise rule.

The other two tasks retain their original-text baselines and support paired comparisons. Medication extraction scores $0.371\pm0.182$ on unmodified text. Under the precise rule, surrogate replacement yields $0.364\pm0.189$, with a paired difference of $-0.008\pm0.178$ that is consistent with zero ($q=0.80$). Under the permissive rule, performance drops to $0.346\pm0.181$, corresponding to a paired difference of $-0.020\pm0.182$ ($q=5.6\times10^{-3}$). Section classification scores $0.749\pm0.093$ on unmodified text. Surrogates reduce this score by $0.072\pm0.086$, whereas placeholders reduce it by $0.038\pm0.089$; both effects are significant ($q=1.6\times10^{-37}$ and $q=2.6\times10^{-14}$; rank-biserial correlations $-0.846$ and $-0.531$). Thus, typed placeholders preserve more utility at the document level, whereas surrogates are required to preserve token-level utility.

\subsection{Leakage}\label{sec:results-leakage}

\begin{table}[t]
  \caption{Leakage from the recommended 13-detector union; all columns concern people.
  \emph{Sens.} is person sensitivity and \emph{Spec.} the share of non-identifier tokens left
  alone: neither is readable without the other. \emph{Exp.} is the risk that a person is still named in a
  given document after processing, \emph{Frequency} the identities named from frequency alone with a public list and,
  bracketed, with the corpus's own distribution (an upper bound, not a risk). \emph{Linkage} is
  the percentage of queries ranking the correct person first, over the complete candidate list,
  against chance $1/\text{gallery}$.}\label{tab:leakage}
  \centering\scriptsize
  \begin{tabular}{@{}l@{\hspace{0.8em}} S[table-format=1.4]@{\hspace{0.8em}} S[table-format=1.4]@{\hspace{0.8em}} S[table-format=1.2]@{\hspace{0.8em}} c@{\hspace{0.8em}} l@{\hspace{0.8em}} l@{\hspace{0.8em}} l@{}}
    \toprule
    Corpus & {Sens.} & {Spec.} & {Exp.\ (\%)} & Frequency & {Context link.} & {Learned link.} & Chance \\
    \midrule
    CARDIO:DE & 0.9998 & 0.8686 & 0.05 & 0 (1) & $0$ & $0.71\pm1.60$ & $1/207$ \\
    Enron     & 0.9906 & 0.6235 & 1.60 & 0 (0) & $0.93\pm0.10$ & $3.94\pm0.30$ & $1/3697$ \\
    \addlinespace
    TAB       & 0.9958 & 0.8504 & 0.83 & 0 (1) & \multicolumn{2}{l}{no links $\rightarrow$ not measurable} & {--} \\
    OntoNotes & 0.9352 & 0.9318 & 5.43 & 0 (1) & \multicolumn{2}{l}{no links $\rightarrow$ not measurable} & {--} \\
    \bottomrule
  \end{tabular}
\end{table}

Table~\ref{tab:leakage} reports the 13-detector union.
This frequency attack with the public name list recovers zero identities by alignment across the
four corpora. The names it did get right were ones the detector missed and the release printed in
clear text, so that count measures defender sensitivity rather than adversarial inference. The corpus distributions, unavailable to a real
attacker, align at most one identity. Raising the abstention threshold from $\varphi=0$ to
$\varphi=0.05$ leaves 11 of 25,677 CARDIO:DE queries and zero elsewhere; alignment $c@1$ is zero at
every threshold. 

Linkage requires cross-document identity and must be read against unmodified text and chance. On
Enron, fixed similarity ranks the correct person first for \SI{71.98}{\percent} of queries on unmodified
text and $0.93\pm0.10$\,\% after replacement. The latter is 34 times the $1/3697$ chance rate
but two orders of magnitude below the unmodified-text rate. Entity-disjoint training raises recovery to
$3.94\pm0.30$\,\%, four times the fixed attack and 146 times chance, so a trained adversary carries
the residual risk. CARDIO:DE has 207 attacker's reference identities; fixed linkage recovery is zero, and
learned linkage $0.71\pm1.60$\,\% over 138 queries against chance $1/207$ is consistent with
guessing.
Across the tested Enron configurations, linkage rises as person sensitivity falls. Agreement
between two detectors lowers sensitivity to 0.8310, raises fixed linkage fivefold to \SI{5.09}{\percent} and
increases clear-text names from 152 to 1,816; intersection yields 0.0443 sensitivity and \SI{25.51}{\percent}
linkage.

The third attack assigns different risks to the replacement forms. Ranking one marked mention against ten candidates, the model falls from 0.995 on unmodified TAB to 0.110 with surrogates, near 0.100 chance, but reaches 0.340 with typed placeholders; OntoNotes (0.290 versus 0.630) orders likewise. Only CARDIO:DE (0.085 versus 0.120) stays at chance level for both categories. This is consistent with source-text recall, not stronger surrogate protection: a neutral placeholder permits retrieval of a memorized person, whereas a surrogate supplies a conflicting name and suppresses that candidate. Accordingly, C is highest on widely replicated OntoNotes, intermediate on public TAB, and at chance on restricted CARDIO:DE. Yet both forms replace identical spans. Replacing each surrogate with \texttt{[PERSON]} reproduces C character for character on all 1,268 TAB and 5,994 OntoNotes documents. 

Sensitivity alone omits exposure units; each denominator answers a different question. The
13-detector union leaves one CARDIO:DE person mention of 2,800 in clear text, in one letter of 400
and for one patient of 270; over every identifier type, \SI{9.5}{\percent} of letters carry
something and \SI{13.3}{\percent} of patients do. TAB leaves a name in 47 of 1,268 judgments and
OntoNotes in 790 of 5,994 documents. On Enron the risk that a person is still named in a given
message is \SI{1.6}{\percent}, while \SI{5.9}{\percent} of messages carry an identifier of some
type.

\label{sec:results-stability}%
The key normaliser both merges and splits identities: on Enron, ORGANISATION surrogates collide at
\SI{17.3}{\percent} and drift at \SI{48.6}{\percent}, and PERSON drift reaches \SI{37.7}{\percent}
against \SI{5.0}{\percent} on CARDIO:DE. The lower CARDIO:DE figure reflects constructed recurrence
with fewer natural variants rather than generalisation. Drift is a utility cost as well as a mapping
failure, because it breaks the longitudinal link that stable pseudonyms exist to preserve. It is a
property of the key normaliser, and a stricter one reduces it at the price of more collisions.

\section{Discussion}\label{sec:discussion}

For releases prioritising leakage reduction, the measurements support the large union, whose
CARDIO:DE specificity is 0.8686. Requiring detector agreement trades sensitivity for specificity.
On Enron, two votes lower person sensitivity from 0.9906 to 0.8310 and raise
fixed linkage fivefold. Across these tested configurations, lower sensitivity coincides with greater
name exposure and linkage.

Attack results depend on adversarial knowledge and denominator. The public prior measures this
frequency attack with one available source; its contrast with the corpus-internal upper bound is the
frequency result. Corpus-level scores should accompany document, case and language results.

For TAB and OntoNotes, a scheme-aware adversary can derive the placeholder release from the surrogate
release. Replacement form still affects utility. Telling the ranker the scheme raises recovery from
0.110 to 0.165 on unchanged text. In the tested tasks, surrogates preserve token-level performance
and placeholders more document-level utility.

The prediction that sensitivity moves frequency alignment and linkage in opposite directions is
refuted for these attacks, priors, corpora and ensemble. Better detection was expected to starve
context linkage while enlarging the population available for frequency alignment. Frequency
alignment recovered zero identities at every tested sensitivity with these priors, while both linkage
attacks declined as sensitivity rose. In these experiments, missed identifiers limited both
measured attacks.

This study has several limitations. Linkage is measurable only on Enron and CARDIO:DE because TAB and
OntoNotes link mentions only within documents. People recur there, so the reported linkage values
understate rather than bound their risk. CARDIO:DE names were inserted into an already de-identified
release, making its frequency distribution synthetic and its clinical realism partial. Linkage
barely works there even before replacement, reaching \SI{2.73}{\percent} on unmodified text:
\SI{56.2}{\percent} of its people appear in a single document, so no profile can be built for them
from others, and discharge letters put much the same words around everyone, leaving two different
profiles a mean cosine of \num{0.305} apart against \num{0.064} on Enron. Its near-zero figures
after replacement therefore describe the corpus more than the protection. Dates were passed unaltered
through our design, yet linkage remained low in our experimentation. The candidate-ranking attack presents ten names and
therefore represents an adversary with a shortlist, not population search. 
\section{Conclusion}\label{sec:conclusion}

Safe sharing of clinical data needs both modalities protected: a guarded image does not compensate
for a report that names the patient. This study measures the text half; the image half is a separate
literature.
Artificial intelligence makes both possible at scale, but its failures must be measured where data
are released. For text we priced detection, utility and re-identification on the same documents. A
large union ensemble replaces almost every person mention and holds both attacks near their chance
rates. Detector precision governs
utility more than the choice between surrogates and placeholders, and sensitivity understates
exposure per document, case and language. There is therefore no single anonymisation score: a
release should state its operating point, per-language errors, surviving cases, attacker knowledge
and stability. Our operating points and measurements are released with acceptance of this paper.


\bibliographystyle{splncs04}
\bibliography{references}

\begin{thebibliography}{10}
\providecommand{\url}[1]{\texttt{#1}}
\providecommand{\urlprefix}{URL }
\providecommand{\doi}[1]{https://doi.org/#1}

\bibitem{akbik2019flair}
Akbik, A., Bergmann, T., Blythe, D., Rasul, K., Schweter, S., Vollgraf, R.:
  {FLAIR}: An easy-to-use framework for state-of-the-art {NLP}. In: Proceedings
  of the 2019 Conference of the North American Chapter of the Association for
  Computational Linguistics (Demonstrations). pp. 54--59 (2019).
  \doi{10.18653/v1/N19-4010}

\bibitem{berg2020impact}
Berg, H., Henriksson, A., Dalianis, H.: The impact of de-identification on
  downstream named entity recognition in clinical text. In: Proceedings of the
  11th International Workshop on Health Text Mining and Information Analysis.
  pp. 1--11. Association for Computational Linguistics (2020).
  \doi{10.18653/v1/2020.louhi-1.1}

\bibitem{bindschaedler2018cryptanalysis}
Bindschaedler, V., Grubbs, P., Cash, D., Ristenpart, T., Shmatikov, V.: The tao
  of inference in privacy-protected databases. Proceedings of the VLDB
  Endowment  \textbf{11}(11),  1715--1728 (2018).
  \doi{10.14778/3236187.3236217}

\bibitem{carrell2013hiding}
Carrell, D., Malin, B., Aberdeen, J., Bayer, S., Clark, C., Wellner, B.,
  Hirschman, L.: Hiding in plain sight: use of realistic surrogates to reduce
  exposure of protected health information in clinical text. Journal of the
  American Medical Informatics Association  \textbf{20}(2),  342--348 (2013).
  \doi{10.1136/amiajnl-2012-001034}

\bibitem{carrell2019parrot}
Carrell, D.S., Cronkite, D.J., Li, M.R., Nyemba, S., Malin, B.A., Aberdeen,
  J.S., Hirschman, L.: The machine giveth and the machine taketh away: a parrot
  attack on clinical text deidentified with hiding in plain sight. Journal of
  the American Medical Informatics Association  \textbf{26}(12),  1536--1544
  (2019). \doi{10.1093/jamia/ocz114}

\bibitem{chambon2023deid}
Chambon, P.J., Wu, C., Steinkamp, J.M., Adleberg, J., Cook, T.S., Langlotz,
  C.P.: Automated deidentification of radiology reports combining transformer
  and ``hide in plain sight'' rule-based methods. Journal of the American
  Medical Informatics Association  \textbf{30}(2),  318--328 (2023).
  \doi{10.1093/jamia/ocac219}

\bibitem{conneau2020xlmr}
Conneau, A., Khandelwal, K., Goyal, N., Chaudhary, V., Wenzek, G., Guzm\'{a}n,
  F., Grave, E., Ott, M., Zettlemoyer, L., Stoyanov, V.: Unsupervised
  cross-lingual representation learning at scale. In: Proceedings of the 58th
  Annual Meeting of the Association for Computational Linguistics. pp.
  8440--8451 (2020). \doi{10.18653/v1/2020.acl-main.747}

\bibitem{dankar2012estimating}
Dankar, F.K., El~Emam, K., Neisa, A., Roffey, T.: Estimating the
  re-identification risk of clinical data sets. BMC Medical Informatics and
  Decision Making  \textbf{12}(1), ~66 (2012). \doi{10.1186/1472-6947-12-66}

\bibitem{din25237}
{DIN Deutsches Institut f{\"u}r Normung e.\,V.}: {DIN EN ISO 25237:2026-06},
  {Medizinische Informatik} --- {Pseudonymisierung} ({ISO}/{DIS} 25237:2026);
  {Deutsche und Englische Fassung} {prEN} {ISO} 25237:2026. Draft standard, DIN
  Media GmbH (2026). \doi{10.31030/3696721}

\bibitem{eder2019emails}
Eder, E., Krieg-Holz, U., Hahn, U.: De-identification of emails: Pseudonymizing
  privacy-sensitive data in a {G}erman email corpus. In: Proceedings of the
  International Conference on Recent Advances in Natural Language Processing
  (RANLP 2019). pp. 259--269. INCOMA Ltd., Varna, Bulgaria (Sep 2019).
  \doi{10.26615/978-954-452-056-4_030}

\bibitem{enisa2021pseudonymisation}
{European Union Agency for Cybersecurity}: Data pseudonymisation: Advanced
  techniques and use cases. Tech. rep., ENISA (2021),
  \url{https://www.enisa.europa.eu/publications/data-pseudonymisation-advanced-techniques-and-use-cases}

\bibitem{gemma2026gemma4}
{Gemma Team}: Gemma 4 technical report (2026), arXiv:2607.02770

\bibitem{hou2025llama}
Hou, Y., Bert, C., Gomaa, A., Lahmer, G., Höfler, D., Weissmann, T., Voigt,
  R., Schubert, P., Schmitter, C., Depardon, A., Semrau, S., Maier, A.,
  Fietkau, R., Huang, Y., Putz, F.: Fine-tuning a local {LLaMA-3} large
  language model for automated privacy-preserving physician letter generation
  in radiation oncology. Frontiers in Artificial Intelligence  \textbf{7},
  1493716 (2025). \doi{10.3389/frai.2024.1493716}

\bibitem{granite41}
{IBM Granite}: granite-4.1-3b (2026), model card,
  \url{https://huggingface.co/ibm-granite/granite-4.1-3b}; no technical report
  published

\bibitem{jha2026piibench}
Jha, P.: {PIIBench}: A unified multi-source benchmark corpus for personally
  identifiable information detection (2026). \doi{10.5281/zenodo.19533263}

\bibitem{johnson2020deid}
Johnson, A.E.W., Bulgarelli, L., Pollard, T.J.: Deidentification of free-text
  medical records using pre-trained bidirectional transformers. In: Proceedings
  of the ACM Conference on Health, Inference, and Learning. pp. 214--221
  (2020). \doi{10.1145/3368555.3384455}

\bibitem{kuzu2013practical}
Kuzu, M., Kantarcioglu, M., Durham, E.A., Toth, C., Malin, B.: A practical
  approach to achieve private medical record linkage in light of public
  resources. Journal of the American Medical Informatics Association
  \textbf{20}(2),  285--292 (2013). \doi{10.1136/amiajnl-2012-000917}

\bibitem{microsoft2025phi4mini}
{Microsoft}: {Phi-4-Mini} technical report: Compact yet powerful multimodal
  language models via mixture-of-{LoRA}s (2025), arXiv:2503.01743

\bibitem{mistral2025magistral}
{Mistral AI}: Magistral (2025), arXiv:2506.10910

\bibitem{mistralsmall32}
{Mistral AI}: {Mistral-Small-3.2-24B-Instruct-2506} (2025), model card; run as
  the FP8 republication
  \url{https://huggingface.co/RedHatAI/Mistral-Small-3.2-24B-Instruct-2506-FP8}

\bibitem{mozes2021nointruder}
Mozes, M., Kleinberg, B.: No intruder, no validity: Evaluation criteria for
  privacy-preserving text anonymization (2021), arXiv:2103.09263

\bibitem{narayanan2008robust}
Narayanan, A., Shmatikov, V.: Robust de-anonymization of large sparse datasets.
  In: 2008 IEEE Symposium on Security and Privacy (sp 2008). pp. 111--125
  (2008). \doi{10.1109/SP.2008.33}

\bibitem{naveed2015inference}
Naveed, M., Kamara, S., Wright, C.V.: Inference attacks on property-preserving
  encrypted databases. In: Proceedings of the 22nd ACM SIGSAC Conference on
  Computer and Communications Security. pp. 644--655 (2015).
  \doi{10.1145/2810103.2813651}

\bibitem{niedermeyer2014cryptanalysis}
Niedermeyer, F., Steinmetzer, S., Kroll, M., Schnell, R.: Cryptanalysis of
  basic bloom filters used for privacy preserving record linkage. Journal of
  Privacy and Confidentiality  \textbf{6}(2) (2014).
  \doi{10.29012/jpc.v6i2.640}

\bibitem{noever2020enron}
Noever, D.: The {Enron} corpus: Where the email bodies are buried? (2020),
  arXiv:2001.10374

\bibitem{openai2025gptoss}
{OpenAI}: gpt-oss-120b \& gpt-oss-20b model card (2025), arXiv:2508.10925

\bibitem{osborne2022bratsynthetic}
Osborne, J.D., Trotter, A., O'Leary, T., Coffee, C., Cochran, M.D.,
  Mansilla-Gonzalez, L., Nadimpalli, A., McAnnally, A., Almudaifer, A.I.,
  Curtis, J.R., Aly, S.M., Kennedy, R.E.: A {Markov} chain replacement strategy
  for surrogate identifiers: Minimizing re-identification risk while preserving
  text reuse. Electronics  \textbf{14}(19), ~3945 (2025).
  \doi{10.3390/electronics14193945}, earlier as arXiv:2210.16125,
  {BRATsynthetic}

\bibitem{packhauser2022reid}
Packh{\"a}user, K., G{\"u}ndel, S., M{\"u}nster, N., Syben, C., Christlein, V.,
  Maier, A.: Deep learning-based patient re-identification is able to exploit
  the biometric nature of medical chest {X}-ray data. Scientific Reports
  \textbf{12}(1),  14851 (2022). \doi{10.1038/s41598-022-19045-3}

\bibitem{packhauser2023anonymization}
Packhäuser, K., Gündel, S., Thamm, F., Denzinger, F., Maier, A.: Deep
  learning-based anonymization of chest radiographs: A utility-preserving
  measure for patient privacy. In: Medical Image Computing and Computer
  Assisted Intervention --- {MICCAI} 2023. pp. 262--272. Lecture Notes in
  Computer Science (2023). \doi{10.1007/978-3-031-43898-1_26}

\bibitem{penas2011simple}
Pe\~{n}as, A., Rodrigo, A.: A simple measure to assess non-response. In:
  Proceedings of the 49th Annual Meeting of the Association for Computational
  Linguistics. pp. 1415--1424 (2011), \url{https://aclanthology.org/P11-1142/}

\bibitem{pilan2022tab}
Pil{\'a}n, I., Lison, P., {\O}vrelid, L., Papadopoulou, A., S{\'a}nchez, D.,
  Batet, M.: The text anonymization benchmark ({TAB}): A dedicated corpus and
  evaluation framework for text anonymization. Computational Linguistics
  \textbf{48}(4),  1053--1101 (2022). \doi{10.1162/coli_a_00458}

\bibitem{presidio}
{Presidio contributors}: Presidio: an open-source framework for detecting,
  redacting, masking and anonymizing sensitive data (2026), software,
  \url{https://github.com/data-privacy-stack/presidio}, formerly
  \texttt{microsoft/presidio}; accessed 2026-09-25

\bibitem{rocher2019estimating}
Rocher, L., Hendrickx, J.M., de~Montjoye, Y.A.: Estimating the success of
  re-identifications in incomplete datasets using generative models. Nature
  Communications  \textbf{10}(1), ~3069 (2019).
  \doi{10.1038/s41467-019-10933-3}

\bibitem{arasteh2024speaker}
Tayebi~Arasteh, S., Arias-Vergara, T., Pérez-Toro, P.A., Weise, T.,
  Packhäuser, K., Schuster, M., Noeth, E., Maier, A., Yang, S.H.: Addressing
  challenges in speaker anonymization to maintain utility while ensuring
  privacy of pathological speech. Communications Medicine  \textbf{4}, ~182
  (2024). \doi{10.1038/s43856-024-00609-5}

\bibitem{trienes2024infolossqa}
Trienes, J., Joseph, S., Schl{\"o}tterer, J., Seifert, C., Lo, K., Xu, W.,
  Wallace, B.C., Li, J.J.: {InfoLossQA}: Characterizing and recovering
  information loss in text simplification. In: Proceedings of the 62nd Annual
  Meeting of the Association for Computational Linguistics (Volume 1: Long
  Papers). pp. 4263--4294. Association for Computational Linguistics, Bangkok,
  Thailand (2024). \doi{10.18653/v1/2024.acl-long.234}

\bibitem{uppala2026privacyfilter}
Uppala, R.: {OpenAI} privacy filter: A cross-lingual, cross-domain {PII}
  evaluation across 32 benchmarks (2026), arXiv:2608.02616

\bibitem{vakili2022utility}
Vakili, T., Dalianis, H.: Utility preservation of clinical text after
  de-identification. In: Proceedings of the 21st Workshop on Biomedical
  Language Processing. pp. 383--388. Association for Computational Linguistics
  (2022). \doi{10.18653/v1/2022.bionlp-1.38}

\bibitem{vats2026redact}
Vats, G., Agrawal, A., Singhal, S., Dash, A., Selvaraj, P., Jhawar, V., Chenna,
  R.P., Bharadwaj, Y.M.G.: {REDACT}: A systematically controlled multilingual
  benchmark for personal information detection (2026), arXiv:2606.19881

\bibitem{vatsalan2014taxonomy}
Vatsalan, D., Christen, P., O'Keefe, C.M., Verykios, V.S.: An evaluation
  framework for privacy-preserving record linkage. Journal of Privacy and
  Confidentiality  \textbf{6}(1) (2014). \doi{10.29012/jpc.v6i1.636}

\bibitem{vidanage2022survey}
Vidanage, A., Ranbaduge, T., Christen, P., Schnell, R.: A taxonomy of attacks
  on privacy-preserving record linkage. Journal of Privacy and Confidentiality
  \textbf{12}(1) (2022). \doi{10.29012/jpc.764}

\bibitem{qwen2025qwen3}
Yang, A., et~al.: Qwen3 technical report (2025), arXiv:2505.09388; the family
  report, no separate report exists for the \texttt{Qwen3.6-35B-A3B} checkpoint
  run here

\bibitem{zaratiana2024gliner}
Zaratiana, U., Tomeh, N., Holat, P., Charnois, T.: {GLiNER}: Generalist model
  for named entity recognition using bidirectional transformer. In: Proceedings
  of the 2024 Conference of the North American Chapter of the Association for
  Computational Linguistics: Human Language Technologies (Volume 1: Long
  Papers). pp. 5364--5376 (2024). \doi{10.18653/v1/2024.naacl-long.300}

\end{thebibliography}

\end{document}